\pdfoutput=1

\documentclass[11pt]{article}

\usepackage[]{acl}

\usepackage{times}
\usepackage{latexsym}

\usepackage[T1]{fontenc}

\usepackage[utf8]{inputenc}

\usepackage{microtype}
\usepackage{graphicx}
\usepackage{multirow}
\usepackage{graphicx}
\usepackage{booktabs}
\usepackage{amsthm}
\usepackage{amsmath}
\usepackage{hyperref}

\title{Semantic Matching from Different Perspectives}

\author{\\
Weijie Liu, 
Tao Zhu, 
Weiquan Mao, 
Zhe Zhao,
Weigang Guo,
Xuefeng Yang
and Qi Ju\textsuperscript{*}\\ 
Tencent Research, Beijing, China\\
\small{dataliu@pku.edu.cn, \{mardozhu, weiquanmao, nlpzhezhao, jimwgguo, ryanxfyang, damonju\}@tencent.com}\\
}

\begin{document}
\maketitle
\begin{abstract}

In this paper, we pay attention to the issue which is usually overlooked, i.e., \textit{similarity should be determined from different perspectives}. To explore this issue, we release a Multi-Perspective Text Similarity (MPTS) dataset, in which sentence similarities are labeled from twelve perspectives.
Furthermore, we conduct a series of experimental analysis on this task by retrofitting some famous text matching models. Finally, we obtain several conclusions and baseline models, laying the foundation for the following investigation of this issue. The dataset and code are publicly available at Github\footnote{\url{https://github.com/autoliuweijie/MPTS}}.

\end{abstract}

\section{Introduction}

Text similarity matching is a crucial technology in search engines and recommendation systems, which is leveraged to calculate the similarity score between two texts and recall the most similar query from a large number of candidate texts \cite{li2016deep}. In the earlier system, the similarity is measured based on the overlap of terms, e.g., TF-IDF \cite{ramos2003using}, BM25 \cite{robertson2009probabilistic}, etc. In recent years, the community has focused more on the semantic similarity calculated by neural network models.

The semantic retrieval system usually consists of two modules, namely recall and rerank. In the recall module, the most commonly used mode is called bi-encoder, which embeds texts as vectors through an encoder model (e.g., SBERT \cite{reimers-2019-sentence-bert}, BERT-flow \cite{li2020bertflow} and SimCSE \cite{gao2021simcse}), and uses a vector search engine (e.g., FAISS \cite{JDH17}) to recall the nearest neighbor vector for the query text. For rerank module, the candidate text and query text are fed into a cross-encoder classifier to determine whether they are similar (RE2 \cite{yang2019simple}, HCAN \cite{rao2019bridging}).

In order to provide a standard benchmark to compare among various similar matching models, there are many public tasks/datasets available. SemEval STS Task \cite{cer2017semeval} released 8628 sentence pairs, and their similarity is represented by scores between 0.0 and 5.0. Quora Question Pairs (QQP) \cite{iyer2017first} is a dataset containing 400k question-question pairs, labeled with 0/1 tag to indicate whether these pairs are similar. Natural Language Inference (NLI) \cite{bowman2015large, MNLI2018, conneau2018xnli} is a task of determining the inference relation (entailment, contradiction or neutral) between two texts.

\begin{table}[t]
\scriptsize
\centering
\begin{tabular}{p{40pt}p{155pt}}
\toprule
\textbf{Field} & \textbf{Content / Perspectives}\\
\midrule
Sentence A     & Star Wars is very exciting, I want to watch it again. \\ \midrule
Sentence B     & I fell asleep when watching Avatar, a bit boring. \\ \midrule
Similar in     & {Themes, Genre} \\ \midrule
Not Similar in & {Litera, Emotion}\\
\bottomrule
\end{tabular}
\caption{An Example of sentences that draw conflicting conclusions from different perspectives.}
\label{tab:similarity_example}
\end{table}

Although many similarity matching models or datasets have been released, the definition of similarity is still unclear, leading to conflicting conclusions. For example, some people think that Sentence A and B in Table \ref{tab:similarity_example} are similar because their themes are both about Sci-fi movies. However, some people hold the opposite view because of their different emotions and literals. There are many perspectives and dimensions to determine similarity and none of them are necessarily right or wrong, thus causing trouble in defining similarity.

Therefore, in this paper, we would like to pay attention to the issue that are usually ignored, i.e., \textit{similarity should be determined from different perspectives}. Based on this, we first release a MPTS dataset in Section \ref{sec:dataset}. Next, we retrofit some baseline models to adapt to the MPTS in Section \ref{sec:method}, and conduct experimental analysis in Section \ref{sec:experiment}. Finally, valuable conclusions are drawn in Section \ref{sec:conclusion}.

The main contributions of this paper can be summarized as follows:
\begin{itemize}
\item We propose a new point that \textit{similarity should be determined from different perspectives.}
\item We build and release the first Multi-Perspective Text Similarity (MPTS) dataset.
\item A series of baseline models are proposed and analyzed in this paper.
\end{itemize}

\section{Dataset}
\label{sec:dataset}


\subsection{Source}

The text samples of MPTS comes from the plot summaries in the Internet Movie Database (IMDB)\footnote{\url{https://www.imdb.com/}}, labeled with one or more genres \cite{read2010scalable}. We take genres as different perspectives, and then pair the summaries in pairs. If two summaries in a pair have the same genre label, they are considered similar from this particular perspective, otherwise, they are not similar. Table \ref{tab:example} gives a pair example, where the samples are similar in some perspectives, but not similar in other perspectives.

\begin{table}[htb]
\scriptsize
\centering
\begin{tabular}{p{40pt}p{155pt}}
\toprule
\textbf{Field} & \textbf{Content / Perspectives}\\
\midrule
Sentence A     & A magician from a faraway land reveals to Iznogoud a new magic trick: a hopscotch that has the power to turn anyone who jumps on the last square back into a kid. \\ \midrule
Sentence B     & As the dragon slayers are drained of their power in the dungeons, the king activates Code ETD and starts an unexpected rebellion. \\ \midrule
Similar in     & {Adventure, Animation, Comedy, Fantasy} \\ \midrule
Not Similar in & {Action, Crime, Drama, Family, Mystery, Romance, Sci-Fi, Thriller}\\
\bottomrule
\end{tabular}
\caption{An example of pairs in MPTS.}
\label{tab:example}
\end{table}

\subsection{Statistics}

To avoid duplication, each sample could be paired only once. Finally, we got a total of 12,734 pairs with 12 perspectives. All pairs are split into train, dev and test set, which contain 10k, 734, and 2k pairs, respectively. Refer to Table \ref{tab:statistics} for details.

\begin{table}[htb]
\scriptsize
\centering
\begin{tabular}{lc}
\toprule
\textbf{Perspective} & \textbf{\begin{tabular}[c]{@{}c@{}}\# Pairs \\ train$\setminus$dev$\setminus$test\end{tabular}}\\
\midrule
\verb|Action| & {3183$\setminus$206$\setminus$614} \\
\verb|Adventure| & {2472$\setminus$179$\setminus$461} \\
\verb|Animation| & {2066$\setminus$164$\setminus$393} \\ 
\verb|Comedy| & {3050$\setminus$251$\setminus$579} \\ 
\verb|Crime| & {3398$\setminus$244$\setminus$708} \\
\verb|Drama| & {6169$\setminus$449$\setminus$1251}  \\ 
\bottomrule
\end{tabular}
\begin{tabular}{lc}
\toprule
\textbf{Perspective} & \textbf{\begin{tabular}[c]{@{}c@{}}\# Pairs \\ train/dev/test\end{tabular}}\\
\midrule
\verb|Family| & {2308$\setminus$167$\setminus$462} \\ 
\verb|Fantasy| & {1832$\setminus$122$\setminus$360} \\ 
\verb|Mystery| & {3306$\setminus$236$\setminus$690} \\ 
\verb|Romance| & {1505$\setminus$129$\setminus$314} \\ 
\verb|Sci-Fi| & {2281$\setminus$169$\setminus$459} \\ 
\verb|Thriller| & {2188$\setminus$147$\setminus$437} \\ 
\bottomrule
\end{tabular}
\caption{The number of pairs in different perspectives.}
\label{tab:statistics}
\end{table}

\section{Method}
\label{sec:method}

In the industry scenario, there are two modes to match text pair similarity, i.e., bi-encoder mode for recall scene and cross-encder mode for rerank scene \cite{reimers-2019-sentence-bert}.

\begin{figure}[tb]
\centering
\includegraphics[width=0.7\columnwidth]{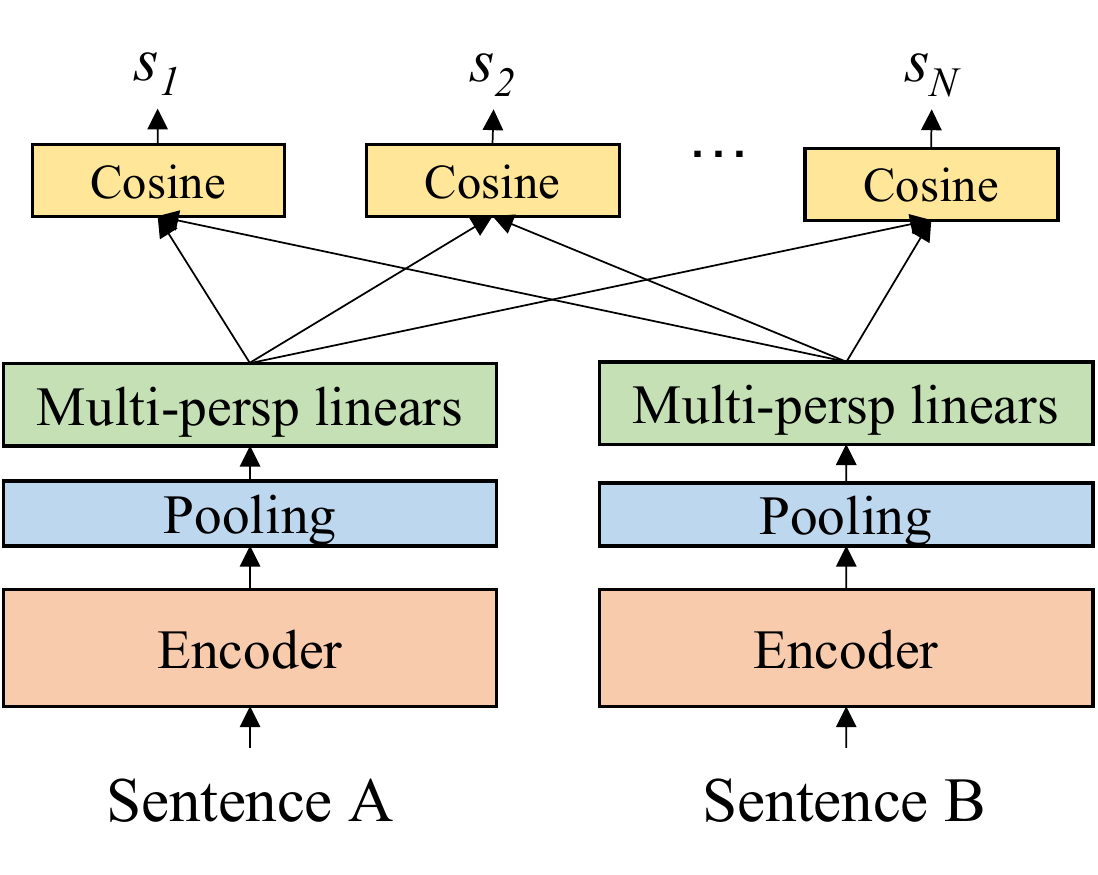}
\caption{Bi-encoder mode for text pair similarity matching in retrieve scene.}
\label{fig:bi-encoder}
\end{figure}

\begin{figure}[tb]
\centering
\includegraphics[width=0.55\columnwidth]{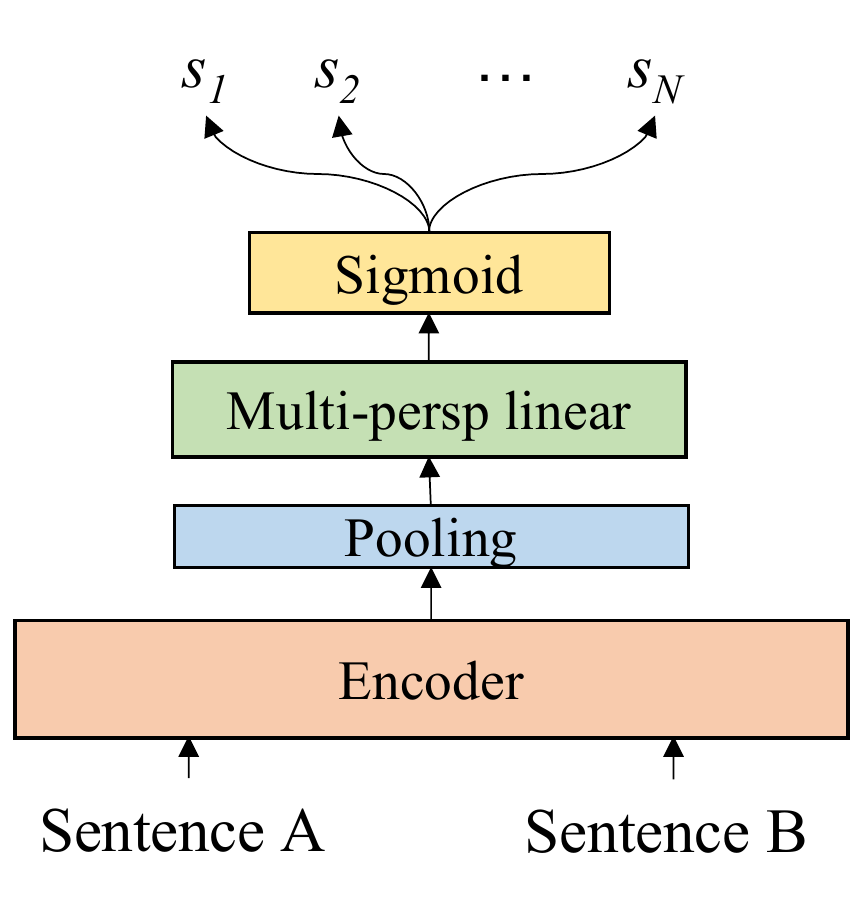}
\caption{Cross-encoder mode for text pair similarity matching in rerank scene.}
\label{fig:cross-encoder}
\end{figure}

\begin{table*}[ht]
\scriptsize
\begin{tabular}{l|l|l|l|l|l|l|l}
\toprule
\textbf{Encoder $\setminus$ Perspective} & \multicolumn{1}{c|}{\textbf{Action}} & \multicolumn{1}{c|}{\textbf{Comedy}} & \multicolumn{1}{c|}{\textbf{Drama}} & \multicolumn{1}{c|}{\textbf{Mystery}} & \multicolumn{1}{c|}{\textbf{Sci-Fi}} & \multicolumn{1}{c|}{\textbf{Thriller}} & \multicolumn{1}{c}{\textbf{W. Avg.}} \\ \midrule
\verb|BERT-base|            & 44.5/85.3/58.5   & 42.9/86.7/57.4   & 74.8/93.4/83.1   & 48.5/90.2/63.1   & 36.0/89.3/51.3   & 32.2/83.7/46.5   & 46.2/89.8/59.8               \\ 
\verb|RoBERTa-base|         & 38.8/77.0/51.6   & 37.3/76.6/50.2   & 73.4/95.0/82.8   & 45.3/87.3/59.6   & 30.7/74.9/43.6   & 25.4/62.4/36.1   & 42.6/82.6/55.2               \\ 
\verb|SBERT-base|           & 37.3/86.1/52.0   & 38.0/87.0/52.9   & 70.4/96.3/81.3   & 42.8/89.8/58.0   & 29.7/84.1/43.9   & 24.8/75.7/37.4   & 41.0/88.8/54.9               \\ 
\verb|SimCSE-BERT-base|     & 43.7/93.1/59.4   & 43.7/93.4/59.6   & 73.0/98.0/83.7   & 49.4/94.3/64.9   & 36.6/93.4/52.6   & 31.8/90.1/47.0   & \underline{46.2/94.2/60.9}   \\ 
\verb|SimCSE-RoBERTa-base|  & 40.8/85.5/55.3   & 39.2/84.9/53.6   & 71.8/95.8/82.1   & 47.1/90.7/62.0   & 33.9/84.1/48.3   & 29.6/79.8/43.2   & 44.0/88.2/57.7               \\ 
\midrule
\verb|BERT-large|           & 44.4/86.4/58.7   & 42.1/88.6/57.1   & 74.0/95.2/83.2   & 49.5/90.0/63.9   & 35.7/89.7/51.0   & 31.2/79.4/44.8   & 45.7/89.4/59.4               \\ 
\verb|RoBERTa-large|        & 38.3/79.6/51.8   & 38.0/82.0/52.0   & 72.1/94.0/81.6   & 45.6/88.4/60.2   & 33.1/83.8/47.5   & 25.6/70.9/37.6   & 42.3/84.6/55.3               \\ 
\verb|SBERT-large|          & 42.6/88.4/57.5   & 39.3/84.6/53.7   & 73.7/94.2/82.7   & 46.4/88.8/61.0   & 34.5/91.0/50.0   & 27.5/76.6/40.5   & 43.9/89.0/57.5               \\ 
\verb|SimCSE-BERT-large|    & 50.5/90.3/64.8   & 48.8/91.0/63.5   & 77.2/96.0/85.6   & 55.1/92.0/68.9   & 42.5/91.2/58.0   & 37.2/86.2/52.0   & \underline{51.6/92.1/65.2}   \\ 
\verb|SimCSE-RoBERTa-large| & 41.8/88.6/56.8   & 39.0/86.8/53.8   & 72.4/94.3/81.9   & 47.8/90.1/62.4   & 34.3/90.2/49.7   & 29.1/81.9/42.9   & 43.9/89.8/57.6               \\ \bottomrule
\end{tabular}
\caption{Evaluation results (precision/recall/F1-score) of the bi-encoder mode using different encoders on MPTS.}
\label{tab:bi-encoder_results}
\end{table*}

\begin{table*}[ht]
\label{tab:comparison}
\scriptsize
\begin{tabular}{l|l|l|l|l|l|l|l}
\toprule
\textbf{Encoder $\setminus$ Perspective} & \multicolumn{1}{c|}{\textbf{Action}} & \multicolumn{1}{c|}{\textbf{Comedy}} & \multicolumn{1}{c|}{\textbf{Drama}} & \multicolumn{1}{c|}{\textbf{Mystery}} & \multicolumn{1}{c|}{\textbf{Sci-Fi}} & \multicolumn{1}{c|}{\textbf{Thriller}} & \multicolumn{1}{c}{\textbf{W. Avg.}} \\ \midrule
\verb|BERT-base|             & 87.9/88.9/88.4  & 89.9/86.8/88.4  & 94.9/93.2/94.1  & 89.6/90.4/90.0   & 88.0/89.7/88.8   & 80.7/84.4/82.5  & \underline{90.3/89.7/90.0} \\ 
\verb|RoBERTa-base|          & 86.9/85.3/86.1  & 88.0/86.1/87.0  & 93.6/93.9/93.7  & 87.0/88.9/88.0   & 87.1/85.6/86.3   & 79.4/76.2/77.8  & 88.3/88.0/88.1 \\ 
\verb|SBERT-base|            & 86.6/89.7/88.1  & 90.4/87.0/88.7  & 94.3/93.8/94.1  & 89.3/91.4/90.4   & 87.7/88.8/88.3   & 79.5/84.4/81.9  & 89.7/90.1/89.8 \\ 
\verb|SimCSE-BERT-base|      & 86.9/88.1/87.5  & 90.0/87.9/88.9  & 95.2/93.5/94.3  & 89.5/91.8/90.7   & 87.8/86.7/87.2   & 82.6/83.7/83.1  & 90.4/89.1/89.7 \\ 
\verb|SimCSE-RoBERTa-base|   & 87.1/85.8/86.4  & 90.7/84.6/87.5  & 93.5/93.6/93.6  & 87.9/89.5/88.7   & 87.5/86.0/86.8   & 78.5/78.0/78.3  & 88.7/88.1/88.4 \\ 
\midrule
\verb|BERT-large|            & 88.9/90.7/89.8  & 90.9/91.8/91.4  & 95.5/94.6/95.1  & 89.1/91.7/90.4   & 90.6/90.8/90.7   & 84.2/83.3/83.7  & 91.2/91.5/91.3  \\ 
\verb|RoBERTa-large|         & 87.8/89.7/88.8  & 89.3/91.3/90.3  & 95.1/94.2/94.7  & 89.3/91.4/90.4   & 87.8/89.7/88.7   & 83.2/83.9/83.6  & 89.7/91.0/90.4  \\ 
\verb|SBERT-large|           & 91.2/87.7/89.4  & 91.6/91.1/91.4  & 95.3/95.2/95.2  & 90.1/91.0/90.5   & 89.6/92.3/90.9   & 82.5/86.5/84.4  & \underline{91.3/91.7/91.5}  \\ 
\verb|SimCSE-BERT-large|     & 89.7/88.7/89.2  & 90.9/89.9/90.4  & 94.6/95.3/94.9  & 88.8/92.0/90.3   & 91.4/91.0/91.2   & 84.3/86.5/85.4  & 91.2/91.5/91.4  \\ 
\verb|SimCSE-RoBERTa-large|  & 88.8/89.4/89.1  & 89.5/88.7/89.1  & 94.7/94.5/94.6  & 90.9/91.3/91.1   & 88.4/91.2/89.8   & 83.9/82.3/83.1  & 90.8/90.8/90.8  \\ 
\bottomrule
\end{tabular}
\caption{Evaluation results (precision/recall/F1-score) of the cross-encoder mode using different encoders on MPTS.}
\label{tab:cross-encoder_results}
\end{table*}

\subsection{Bi-encoder}

The mode of the bi-encoder is shown in Figure \ref{fig:bi-encoder}, where \textit{Sentence A} and \textit{Sentence B} are respectively encoded with two parameter-sharing encoders. The encoder can be BERT \cite{devlin2018bert}, RoBERTa \cite{liu2019roberta}, etc. The pooling type could be selected from \textit{CLS}, \textit{Last-Avg}, \textit{First-Last-Avg}, \textit{Last2-Avg} \cite{su2021whitening}. The multi-persp linears consists of $N$ dense linear layers, which convert the single sentence embedding into $N$ normalized embeddings corresponding to the $N$ perspectives. For the $i (i=1,2, ..., N)$ perspective, the similarity score $s_i \in [-1, 1]$ is the $Cosine$ value calculated from the embeddings of \textit{sentence A} and \textit{sentence B} in the corresponding perspective. When $s_i \geq 0$, it is regarded as similar, otherwise, it is not. When training, the bi-encoder mode adopts the loss function in (\ref{eq:bi-encoder_loss}), 
\begin{equation}
\label{eq:bi-encoder_loss}
\begin{aligned}
loss =& \sum_{i=1}^{N}y_{i}log\frac{e^{s_i/\tau} }{e^{s_i/\tau} + e^{-s_i/\tau}} \\ & + \sum_{i=1}^{N}(1-y_i)log\frac{e^{-s_i/\tau}}{e^{s_i/\tau} + e^{-s_i/\tau}},
\end{aligned}
\end{equation} 
where $y_i=1$ when the pair is similar from the $i$-th perspective, otherwise $y_i=0$. $\tau$ is a hyperparameter called temperature.

\subsection{Cross-encoder}

The mode of the cross-encoder is shown in Figure \ref{fig:cross-encoder}, \textit{Sentence A} and \textit{Sentence B} are sent to the same encoder and pooling layer, and are represented as a $H$-dimensional embedding vector. In this mode, \textit{multi-persp linear} is a dense linear layer of $H\times N$, which converts the $H$-dimensional embedding vector into a $N$-dimensional score vector corresponding to the $N$ perspectives. The similarity score $s_i$ is restricted between $0$ and $1$ with a $Sigmoid$ function. When $s_i \geq 0.5$, it is regarded as similar, otherwise, it is not. For training, the cross-encoder mode uses the loss function in (\ref{eq:cross-encoder_loss}),
\begin{equation}
\label{eq:cross-encoder_loss}
\begin{aligned}
loss = \sum_{i=1}^{N}y_{i}log(s_i) + \sum_{i=1}^{N}(1-y_i)log(1-s_i).
\end{aligned}
\end{equation} 

\section{Experimental results}
\label{sec:experiment}

In this section, we equip the bi-encoder and cross-encoder modes with different sentence representation encoders, and show and analyze some experimental results on MPTS. SimCSE encoders are unsupervised versions \cite{gao2021simcse}, and SBERT encoders are NLI versions \cite{reimers-2019-sentence-bert}. All models are trained on the \textit{train} set for 10 epochs, and then the model with best performance in the \textit{dev} set is selected for evaluation using the \textit{test} set. More training details can be found in Appendix \ref{sec:training_details}.

\subsection{Impact of pooling type}

\begin{figure}[bt]
\centering
\includegraphics[width=0.9\columnwidth]{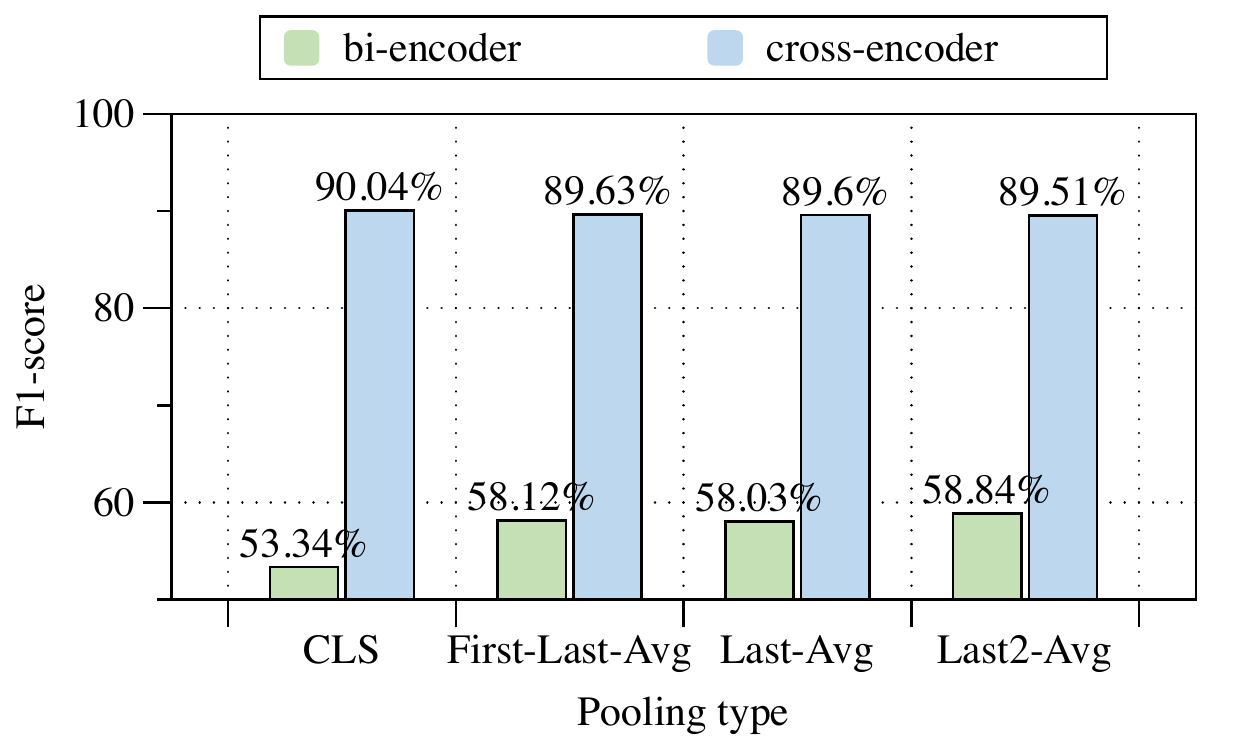}
\caption{MPTS performance of the BERT-base model with different pooling types.}
\label{fig:pooling_type_affect}
\end{figure}

For both bi-encoder and cross-encoder modes, the pooling type is a hyper-parameter that needs to be determined. In Figure \ref{fig:pooling_type_affect}, we show the weighted F1-score on the MPTS of these two modes using a BERT-base \cite{devlin2018bert} encoder with different pooling types. 

It can be observed that for the bi-encoder mode, the performance of \textit{CLS} pooling type is worse than the other three, and these four pooling types have little effect on the cross-encoder. Based on this observation, in this paper, all bi-encoder models use \textit{Last2-Avg}, and cross-encoder models use \textit{CLS}.

\subsection{Impact of temperature}

\begin{figure}[bt]
\centering
\includegraphics[width=0.9\columnwidth]{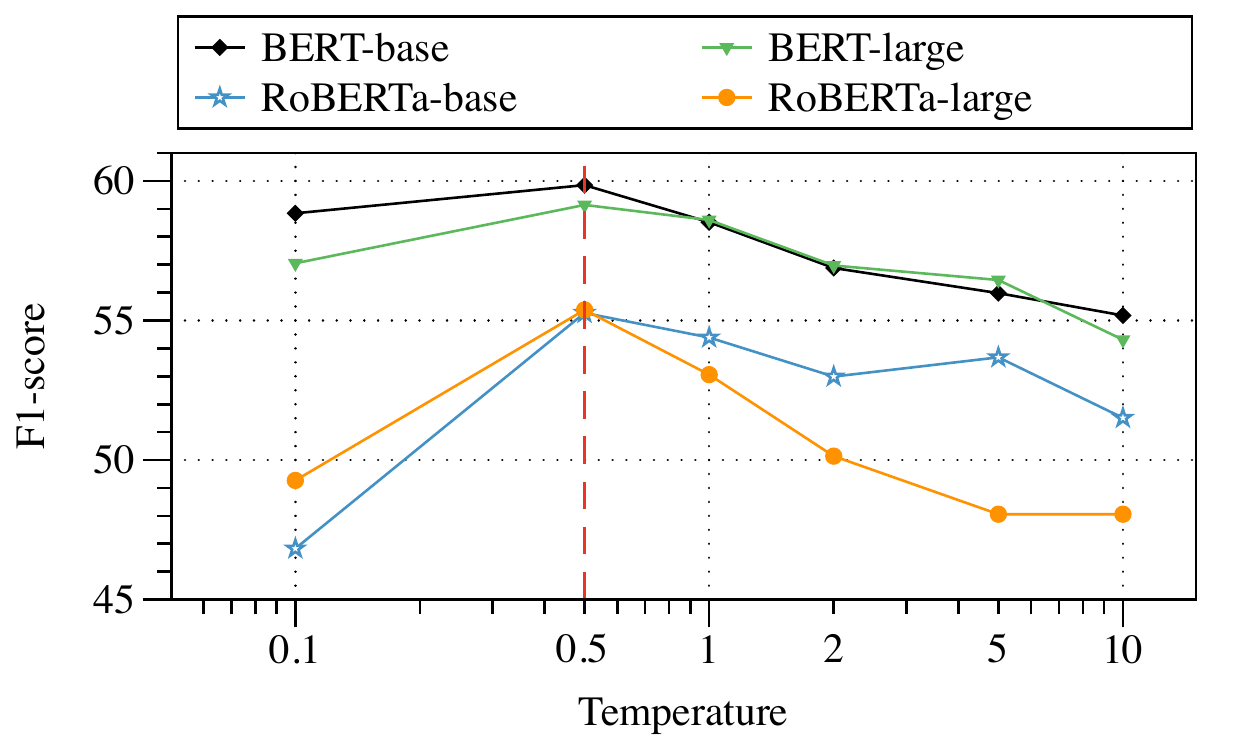}
\caption{MPTS performance of bi-encoder models with different temperature $\tau$ setting.}
\label{fig:temperature_affect}
\end{figure}

The temperature $\tau$ in (\ref{eq:bi-encoder_loss}) is a key hyper-parameter for bi-encoder models. Appropriate $\tau$ will bring high performance, while bad $\tau$ may even make the model not converge. In this experiment, we take BERT-base/large \cite{devlin2018bert} and RoBERT-base/large  \cite{liu2019roberta, zhao2019uer} as examples to explore the impact of temperature $\tau$, and the results are shown in Figure \ref{fig:temperature_affect}.

It can be found that $\tau$ has a great influence on these models, and $\tau=0.5$ is their common peak point. In addition, we found that if $\tau<0.1$ may cause model overflow during the training process, and if $\tau>10$ will make the model not converge. Therefore, $\tau=0.5$ is the default setting for all the bi-encoder models in this paper.

\subsection{Results of bi-encoder}

In Table \ref{tab:bi-encoder_results}, we show the bi-encoder models' evaluation metrics of six representative perspectives selected from the 12 MPTS perspectives, and the W. Avg. is the weighted average score of all 12 perspectives. The metrics of the other six perspectives can be found in Appendix \ref{sec:additional_benchmarks}. It can be seen that SimCSE-Bert-base \cite{gao2021simcse} has the best performance among all base-size encoders, followed by BERT-base\cite{devlin2018bert}. For large-size encoders, SimCSE-Bert-large \cite{gao2021simcse} vastly superior with 65.2\% F1-score. However, compared with the cross-encoder models in Table \ref{tab:cross-encoder_results}, its performance still has plenty of room for improvement.

\subsection{Results of cross-encoder}

Table \ref{tab:cross-encoder_results} shows the evaluation metrics of the cross-encoder models for six representative perspectives, and that of the remaining six perspectives can be found in the Appendix \ref{sec:additional_benchmarks}. Benefited from the deep interaction between sentences in cross-encoder mode, its performance is significantly better than bi-encoder mode, and the performance divergence among different encoders is relatively small. Specifically, the F1-score range of the base-size models is 88\% to 90\%, and that of the large-size model is 90\% to 92\%.

\subsection{Impact of model size}

Table \ref{fig:size_affect} compares the performance of different model sizes, where BERT-tiny/mini/small/medium come from \citet{Turc2019bert} and BERT-base/large come from \citet{devlin2018bert}. It can be seen that no matter which mode it is, the performance is positively correlated with the model size.

\begin{table}[htb]
\centering
\scriptsize
\begin{tabular}{lccc}
\toprule
\multicolumn{1}{l}{\textbf{Encoder}} & \multicolumn{1}{c}{\textbf{Size}} & \multicolumn{1}{c}{\textbf{Bi-encoder}} & \multicolumn{1}{c}{\textbf{Cross-encoder}} \\ \midrule
BERT-tiny    &  4.3M & 52.90\% & 75.94\% \\
BERT-mini    & 11.1M & 53.67\% & 80.76\% \\
BERT-small   & 28.5M & 58.13\% & 85.64\% \\
BERT-medium  & 41.1M & 56.54\% & 87.61\% \\
BERT-base    &  110M & 59.85\% & 90.04\% \\
BERT-large   & 340M  & 59.41\% & 91.38\% \\
\bottomrule
\end{tabular}
\caption{MPTS F1-scores of BERT with different sizes.}
\label{fig:size_affect}
\end{table}

\section{Conclusion}
\label{sec:conclusion}

In this work, we first propose the point that \textit{similarity should be judged from different perspectives}. In order to explore the feasibility of judging similarity from different perspectives, we constructed and released a multi-perspective text similarity dataset (MPTS). After that, a series of experiments were carried out on this dataset with the commonly used modes in industrial scenes, i.e., bi-encoder and cross-encoder. Finally, we draw the following conclusions: 1. By adding a \textit{multi-persp linear} layer, both bi-encoder and cross-encoder modes can be applied to the multi-perspective similarity matching task. 2. The bi-encoder, as an essential mode in retrieval systems, still has plenty of room for improvement in this task. 3. The performance is positively associated with the model size, no matter which mode it is.

\bibliography{main.bib}
\bibliographystyle{acl_natbib}

\appendix

\begin{table*}[t]
\centering
\scriptsize
\begin{tabular}{l|l|l|l|l|l|l}
\toprule
\textbf{Encoder $\setminus$ Perspective} & \multicolumn{1}{c|}{\textbf{Adventure}} & \multicolumn{1}{c|}{\textbf{Animation}} & \multicolumn{1}{c|}{\textbf{Crime}} & \multicolumn{1}{c|}{\textbf{Family}} & \multicolumn{1}{c|}{\textbf{Fantasy}} & \multicolumn{1}{c}{\textbf{Romance}}  \\ \midrule
\verb|BERT-base|            & 35.3/89.1/50.6   & 33.7/94.1/49.7   & 53.1/95.3/68.2   & 36.4/93.2/52.3   & 27.3/84.7/41.3   & 24.2/82.4/37.4                  \\ 
\verb|RoBERTa-base|         & 31.4/78.5/44.9   & 30.9/90.8/46.2   & 50.7/93.9/65.9   & 34.7/87.6/49.7   & 23.4/68.8/34.9   & 19.5/63.0/29.8                  \\ 
\verb|SBERT-base|           & 31.5/89.1/46.6   & 28.2/92.1/43.2   & 48.3/96.1/64.3   & 31.6/92.2/47.1   & 22.3/77.2/34.7   & 20.5/77.7/32.5                  \\ 
\verb|SimCSE-Bert-base|     & 35.7/93.9/51.8   & 33.8/96.1/50.0   & 53.6/98.0/69.3   & 37.5/93.2/53.5   & 27.0/87.2/41.2   & 25.1/87.5/39.1      \\ 
\verb|SimCSE-roberta-base|  & 34.0/85.4/48.6   & 32.9/92.8/48.6   & 51.4/94.9/66.7   & 35.1/87.8/50.2   & 35.1/87.8/50.2   & 23.7/80.5/36.6                  \\ 
\midrule
\verb|BERT-large|           & 33.4/83.5/47.7   & 32.4/93.1/48.1   & 52.4/94.6/67.5   & 37.1/93.9/53.2   & 26.8/80.5/40.2   & 23.1/82.1/36.1                  \\ 
\verb|RoBERTa-large|        & 31.5/81.7/45.5   & 30.1/92.3/45.4   & 50.0/92.6/64.9   & 33.0/86.1/47.7   & 23.3/72.2/35.2   & 18.8/63.0/28.9                  \\ 
\verb|SBERT-large|          & 32.7/88.0/47.7   & 29.9/94.1/45.4   & 50.3/95.6/65.9   & 34.8/91.1/50.4   & 25.5/81.3/38.8   & 21.3/77.3/33.5                  \\ 
\verb|SimCSE-Bert-large|    & 41.8/92.6/57.6   & 39.1/93.8/55.2   & 59.2/95.6/73.1   & 42.8/94.1/58.8   & 33.1/85.5/47.7   & 29.0/86.3/43.5      \\ 
\verb|SimCSE-roberta-large| & 33.0/92.1/48.6   & 30.2/92.8/45.6   & 51.4/94.2/66.5   & 34.0/90.9/49.5   & 24.2/81.1/37.3   & 22.1/81.2/34.7                  \\ 
\bottomrule
\end{tabular}
\caption*{Table A.1: The evaluation results (precision/recall/F1-score) of the bi-encoder models on the remaining six MPTS perspectives.}
\end{table*}

\begin{table*}[t]
\centering
\scriptsize
\begin{tabular}{l|l|l|l|l|l|l}
\toprule
\textbf{Encoder $\setminus$ Perspective} & \multicolumn{1}{c|}{\textbf{Adventure}} & \multicolumn{1}{c|}{\textbf{Animation}} & \multicolumn{1}{c|}{\textbf{Crime}} & \multicolumn{1}{c|}{\textbf{Family}} & \multicolumn{1}{c|}{\textbf{Fantasy}} & \multicolumn{1}{c}{\textbf{Romance}}  \\ \midrule
\verb|BERT-base|            & 85.6/88.2/86.9   & 92.5/88.8/90.6   & 95.7/95.2/95.4   & 91.2/91.9/91.5   & 85.5/82.2/83.8   & 91.7/85.3/88.4   \\ 
\verb|RoBERTa-base|         & 81.6/87.6/84.5   & 90.1/88.8/89.4   & 93.3/94.9/94.1   & 89.8/90.2/90.0   & 79.9/80.8/80.3   & 90.3/80.2/84.9   \\ 
\verb|SBERT-base|           & 85.3/88.5/86.9   & 91.9/90.5/91.2   & 95.4/95.6/95.5   & 89.1/91.9/90.5   & 84.2/83.0/83.6   & 90.9/83.1/86.8   \\ 
\verb|SimCSE-Bert-base|     & 86.0/84.1/85.0   & 89.8/89.5/89.6   & 95.2/95.2/95.2   & 91.1/91.3/91.2   & 89.2/78.6/83.6   & 92.8/83.1/87.7    \\ 
\verb|SimCSE-roberta-base|  & 81.3/87.2/84.1   & 90.4/88.8/89.6   & 93.1/95.6/94.3   & 90.3/90.9/90.6   & 82.3/80.5/81.4   & 90.3/80.2/84.9    \\ 
\midrule
\verb|BERT-large|           & 85.4/90.4/87.8   & 91.2/93.1/92.1   & 96.5/95.4/96.0   & 94.2/92.8/93.5   & 86.7/87.2/86.9   & 92.1/85.3/88.6     \\ 
\verb|RoBERTa-large|        & 83.6/87.4/85.4   & 90.0/92.1/91.0   & 95.1/95.9/95.5   & 88.7/93.9/91.2   & 84.0/85.0/84.5   & 90.4/87.2/88.8     \\ 
\verb|SBERT-large|          & 86.1/90.2/88.1   & 91.3/91.6/91.4   & 94.5/96.3/95.4   & 93.9/94.8/94.4   & 87.0/87.5/87.2   & 93.1/86.6/89.7     \\ 
\verb|SimCSE-Bert-large|    & 86.5/90.4/88.4   & 90.8/93.6/92.2   & 95.8/95.3/95.6   & 94.1/94.1/94.1   & 88.1/84.4/86.2   & 92.5/86.9/89.6     \\ 
\verb|SimCSE-roberta-large| & 87.7/88.2/88.0   & 90.1/93.1/91.6   & 95.4/97.3/96.3   & 90.3/94.8/92.5   & 90.6/80.5/85.2   & 90.5/85.0/87.6     \\ \bottomrule
\end{tabular}
\caption*{Table A.2: The evaluation results (precision/recall/F1-score) of the cross-encoder models on the remaining six MPTS perspectives.}
\end{table*}

\section{Supplementary results}
\label{sec:additional_benchmarks}

Table A.1 and Table A.2 respectively give the evaluation results of the bi-encoder and cross-encoder models in the other six MPTS perspectives.

\section{Training details}
\label{sec:training_details}

We implement all models based on Huggingface's \textit{transformers} package \cite{wolf2020transformers}. The maximum sequence length of the bi-encoder and cross-encoder models is set to 128 and 256, respectively. We use the AdamW optimizer \cite{loshchilov2017decoupled} to train these models for 10 epochs with a batch size of 32, and use the model with the best performance on the \textit{dev} set for finally evaluation. The learning rate is set as 5e-5 and weight decay is 0.01. For bi-encoder models, the temperature is set as 0.5.

\end{document}